\documentclass{article}
\usepackage[T1]{fontenc}
\usepackage{mathptmx}
\usepackage{arxiv}
\usepackage{amsmath,amssymb}
\usepackage{graphicx}
\usepackage{booktabs,array}
\usepackage{caption}
\usepackage{float}
\usepackage{microtype}
\usepackage[hidelinks]{hyperref}
\renewcommand{\thetable}{\Roman{table}}
\renewcommand{\thesection}{\Roman{section}}
\renewcommand{\thesubsection}{\Alph{subsection}}
\makeatletter
\renewcommand{\p@subsection}{\thesection-}
\makeatother
\newcommand{\tanhfpu}{$\tanh$(FPU)}
\renewcommand{\shorttitle}{MulClip: A Table-Free Activation Function for Reservoir Computing}
\renewcommand{\undertitle}{}

\title{MulClip: A Table-Free Activation Function with Monotonic and Folding Regimes for Reservoir Computing on Microcontrollers}
\author{Guren Matsumura, Masato Takemoto, and Shinzo Koyama\\
  \normalfont Panasonic Industry Co., Ltd., Osaka, Japan\\
  \normalfont\small\texttt{matsumura.guren@jp.panasonic.com}}
\date{}

\begin{document}
\raggedbottom
\maketitle

\begin{abstract}
This letter proposes MulClip, a lightweight activation function for reservoir computing (RC) on microcontrollers (MCUs). In a ring reservoir, a representative sparse RC topology, multiply--accumulate operations are few, and evaluating $\tanh$ accounted for 75\% of the state-update time on a Cortex-M4. MulClip introduces curvature into the response using only comparison, absolute value, multiplication, and subtraction, and places its clipping threshold not at the extremum but at the point where the response returns to zero after passing the extremum. As a result, a single function has both a monotonic regime and a nonmonotonic folding regime that increases the expressiveness of a single unit. Which regime is used is determined by the usual hyperparameter search, e.g., over the input scale, so no task-specific threshold tuning is required. The state-update time measured on hardware was 0.34 times that of $\tanh$ and shorter than that of any of the 11 evaluated $\tanh$ approximations. In PC-based evaluation, third-order MulClip achieved a 22\% lower NMSE than $\tanh$ on NARMA-20, where the pre-activations remained in the monotonic regime, and on 8-bit parity it raised the accuracy from 0.76--0.78 for the monotonic-clip variant without folding to 0.97--0.98. For these two tasks, on-device inference yielded nearly the same values as on the PC.
\end{abstract}

\keywords{Activation function \and embedded machine learning \and microcontroller \and reservoir computing}

\section{Introduction}\label{sec:intro}
In recent years, there has been growing demand for running machine-learning inference on embedded devices, for example, for sequential processing of sensor signals \cite{warden}--\cite{cmsis}. However, the microcontrollers (MCUs) widely used in embedded devices have limited computing performance and memory capacity, and it is often difficult to complete inference within the required processing period.

Reservoir computing (RC) \cite{jaeger}, \cite{lukosevicius} is a lightweight approach suited to such environments. RC fixes the reservoir weights and trains only the readout layer. Moreover, sparsifying the reservoir connections reduces the multiply--accumulate operations in the state update. A ring topology, in which connections are restricted to a cycle, achieves performance comparable to that of a randomly connected reservoir while making the number of multiply--accumulate operations roughly proportional to the number of units \cite{rodan}. However, the fewer the multiply--accumulate operations, the larger the share of the activation function in the state-update cost. The standard $\tanh$ is a transcendental function and is expensive relative to multiplication and addition even on MCUs with an FPU \cite{arm}. Indeed, for a ring reservoir on a Cortex-M4, $\tanh$ evaluated on the FPU (hereafter \tanhfpu) accounted for 75\% of the state-update time (Section~\mbox{\ref{sec:cost}}).

Alternatives to $\tanh$ include implementations that approximate $\tanh$ with LUTs, piecewise-linear (PWL) approximations, or rational functions \cite{amin}, \cite{baker}, and functions such as hardtanh, softsign \cite{glorot}, and ReLU \cite{nair}. The former aim to reproduce $\tanh$, so higher fidelity requires more memory or arithmetic. The latter are inexpensive but have fixed shapes; in our measurements, they increased the NARMA-20 error to 1.6--1.9 times that of \tanhfpu. The shape of the activation function strongly affects RC performance \cite{hurley}, and in particular, responses that are nonmonotonic in the input are known to increase the expressiveness of a single unit \cite{noel}. However, most such functions involve transcendental functions and thus suffer from the same problem as $\tanh$ on MCUs.

This letter therefore proposes MulClip, an activation function that, rather than approaching $\tanh$, inexpensively provides a response that $\tanh$ lacks. MulClip introduces curvature using only comparison, absolute value, multiplication, and subtraction, and places the clipping threshold at the point where the output returns to zero, which is determined by the functional form; thus, a single function has both a monotonic regime and a folding regime without task-specific threshold tuning. This is the central contribution of this letter. With this design, both a monotonic operating point and an operating point that exploits folding can be selected solely through the hyperparameter search over, e.g., the input scale, that is routinely performed in RC. Section~\ref{sec:method} presents the design, and Section~\ref{sec:exp} evaluates it in terms of both on-device computational cost and task accuracy.

\section{Proposed Method---MulClip}\label{sec:method}
The cheapest way to provide a bounded nonlinearity is to clip the input; clipping with a threshold of 1 yields hardtanh. MulClip adds a low-order term with a negative coefficient to the clipped input to give the response curvature. MulClip of order $d$ ($d\geq2$) is defined as follows.
\begin{gather}
 x_c=\min\bigl(\max(x,-c_d),\,c_d\bigr),\label{eq:clip}\\
 f(x)=x_c-\alpha_d\,|x_c|^{d-1}x_c,\qquad \alpha_d=\frac{(d-1)^{d-1}}{d^d},\label{eq:f}\\
 c_d=\alpha_d^{-1/(d-1)}=\frac{d^{\,d/(d-1)}}{d-1}.\label{eq:cd}
\end{gather}
The coefficient $\alpha_d$ is chosen such that the unclipped response attains its extremum of 1 at $|x|=d/(d-1)$, and $c_d$ is the point where that response returns to 0 after passing the extremum (for $d=2$, $\alpha_2=1/4$ and $c_2=4$; for $d=3$, $\alpha_3=4/27$ and $c_3=3\sqrt3/2\approx2.60$). Evaluation requires only two comparisons, $d$ multiplications, and one subtraction (plus one absolute value for $d=2$), with no division, transcendental function, or table lookup. The output always lies in $[-1,1]$.

\begin{figure}[!tbp]
 \centering
 \includegraphics[width=0.85\linewidth]{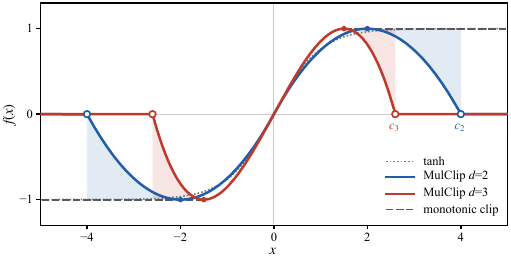}
 \caption{Response of MulClip (blue: $d=2$; red: $d=3$). Small dots mark the extremum at $d/(d-1)$, and open circles mark the clipping threshold $c_d$. Gray dashed lines show the monotonic-clip variant, whose clipping threshold is set at $d/(d-1)$. The dotted line is $\tanh$.}\label{fig:response}
\end{figure}

Fig.~\ref{fig:response} shows the response. Since MulClip is an odd function, we describe the $x\geq0$ side below. For $0\leq x\leq d/(d-1)$, the response increases monotonically from a slope of 1 at the origin until it reaches 1 (monotonic regime). Beyond this point, it decreases and returns to 0 at $c_d$ (folding regime). Beyond $c_d$, the output is 0 (zero regime). The folding regime traverses the same output range $[0,1]$ as the monotonic regime in reverse. If the clipping threshold is smaller than $c_d$, the folding stops partway; if it is larger, a region arises in which the sign of the output is opposite to that of the input.

If the input to the activation function (the pre-activation) stays within the monotonic regime, the output of MulClip coincides with that of the monotonic function whose clipping threshold is set at $d/(d-1)$ (hereafter, the monotonic-clip variant). In RC, hyperparameters such as the input scale are customarily searched for each task \cite{lukosevicius}, and this search determines which regime is used. Therefore, the clipping threshold need not be tuned for each task.

The polynomial part of MulClip itself is not new. For example, the monotonic-clip variant with $d=3$ equals the cubic soft clipper \cite{smith} scaled so that both the slope at the origin and the maximum are 1. MulClip moves this clipping threshold from the extremum to $c_d$, where the output returns to 0. Making the function odd through a sign-multiplied even-order term is shared with the polynomial activation functions of Abe et al.\ \cite{abe}, but those functions maintain a monotonic response through positive coefficients and output saturation. Hurley et al.\ showed that nonmonotonic functions are effective for RC \cite{hurley}, but their function involves transcendental functions. Yoshioka and Tamukoh implemented LUT-based RC on low-cost FPGAs \cite{yoshioka}. The novelty of this letter lies not in the polynomial or the folding response per se, but in making a monotonic regime and a folding regime coexist in a single activation function, without adding a task-specific threshold parameter, through the clipping threshold $c_d$ uniquely determined by the functional form.

\section{Experimental Evaluation}\label{sec:exp}
\subsection{Experimental Setup}\label{sec:setup}
The state-update time was measured by running a 500-unit ring reservoir, in which each unit receives only the state of the preceding unit,
\begin{equation}
 x_i(t+1)=f\bigl(w\,x_{i-1}(t)+v_i u(t)+b_i\bigr)\qquad (x_0=x_N,\ N=500)\label{eq:ring}
\end{equation}
for 230 steps on an STM32 NUCLEO-F446RE (Cortex-M4, single-precision FPU, 84~MHz). Here, $w$ is the shared coupling weight, $v_i$ and $b_i$ are the input weight and bias, and the argument of $f$ is the pre-activation. The state-update time (the time to compute the pre-activations and evaluate the activation function for all units) was measured with the cycle counter (firmware built with \texttt{-O2}). We compared 20 activations: MulClip ($d=2,3$); \tanhfpu, i.e., newlib's \texttt{tanhf} executed on the FPU; 11 $\tanh$ approximations (two rational approximations, five nearest-neighbor LUTs, three interpolated LUTs, and an eight-segment PWL approximation, PWL8); hardtanh; softsign; ReLU; and three polynomial activation functions of Abe et al.\ (F1--F3) \cite{abe} (implementation and optimization details in Appendices~\ref{app:A} and~\ref{app:B}).

Accuracy was evaluated by PC simulation (reservoir state update in float32, readout in float64) on four tasks: NARMA-20 \cite{atiya} (NMSE); 8-bit parity, Parity8 \cite{bertschinger} (accuracy); autonomous prediction of the Mackey--Glass time series \cite{mackey}, MG (valid prediction time, VPT, in steps); and 10-word recognition on Google Speech Commands \cite{speech}, KWS (accuracy). The readout was trained by ridge regression. For each activation, the spectral radius (coupling weight $w$), input scale, and other hyperparameters were searched with 100 trials of Optuna's TPE \cite{optuna}, and we report averages over 50 random seeds separate from those used in the search (Appendix~\ref{app:C}). To isolate the effect of the folding response, the monotonic-clip variant was evaluated with the same procedure. The on-device MulClip states agreed with the PC computation within a maximum absolute error of $2.0\times10^{-6}$, even under conditions in which pre-activations entered the folding and zero regimes. Furthermore, for NARMA-20 and Parity8, we ran inference over the evaluation segment on the device using readouts trained on the PC (five seeds) and obtained nearly the same NMSE and accuracy as on the PC (Appendix~\ref{app:F}).

\subsection{Evaluation Cost}\label{sec:cost}
The state-update time of MulClip was comparable to that of the cheapest functions, which use neither tables nor transcendental functions, and shorter than that of every $\tanh$ approximation. Evaluating \tanhfpu\ required 203.2 cycles per activation and accounted for 75\% of the 1.615-ms state-update time. MulClip required 30.0 ($d=2$) and 31.0 ($d=3$) cycles per activation, giving state-update times of 0.543 and 0.549~ms (0.34 times that of \tanhfpu). These are within 0.06~ms of hardtanh, softsign, and ReLU (0.490--0.531~ms), whereas all 11 $\tanh$ approximations (0.592--0.704~ms) took longer. The state-update times of the functions of Abe et al.\ (F1--F3) were 0.34--0.48 times that of \tanhfpu\ (measured in a separate build; Appendix~\ref{app:B}). The flash usage (text+data) of binaries built with only one activation each exceeded that of the smallest ones, ReLU and softsign, by 40--56~B for MulClip and by 880~B for \tanhfpu. The per-step processing time including the readout was 1.67~ms for \tanhfpu\ and 0.60~ms for MulClip with $d=3$; only \tanhfpu\ exceeds a 1-kHz processing period. In terms of cost, MulClip is thus nearly on par with the cheapest fixed-shape functions, and its difference from them appears in accuracy, as shown in the next subsection.

\subsection{Accuracy Versus State-Update Time}\label{sec:acc}
\begin{figure}[!tbp]
 \centering
 \includegraphics[width=0.80\linewidth]{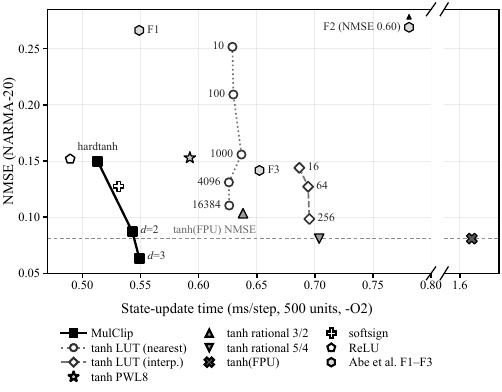}
 \caption{NMSE versus state-update time (500 units) on NARMA-20. Lower left is better. The horizontal axis is broken. The horizontal dashed line indicates the NMSE of \tanhfpu. The connected black squares are, from left to right, hardtanh, MulClip with $d=2$, and MulClip with $d=3$. Numbers beside the LUT points denote the number of entries. Hexagons denote the functions of Abe et al.\ \cite{abe}; F2 (NMSE 0.60) lies outside the vertical-axis range and is plotted at the top edge. The times for the functions of Abe et al.\ were measured in a separate build (Appendix~\ref{app:B}).}\label{fig:narma}
\end{figure}

Fig.~\ref{fig:narma} shows the relationship between NMSE and state-update time on NARMA-20. Adding curvature to hardtanh lowered the NMSE from 0.150 to 0.087 for $d=2$ and to 0.063 for $d=3$; the latter is 22\% below the 0.081 of \tanhfpu. Because the pre-activations never entered the folding regime on NARMA-20 (Section~\mbox{\ref{sec:folding}}), this improvement is attributable to curvature. None of the $\tanh$ approximations achieved a lower NMSE than \tanhfpu, and \tanhfpu\ and all 11 approximations had both longer state-update times and higher NMSEs than MulClip with $d=3$.

\begin{table}[!tbp]
 \centering\small
 \caption{Accuracy (Mean Over 50 Seeds), State-Update Time, and Flash Usage (Relative to \tanhfpu) of Each Activation}\label{tab:main}
 \setlength{\tabcolsep}{4pt}
 \begin{tabular}{lcccccc}
 \toprule
 & NARMA-20 & Parity8 & MG & KWS & State-update & Flash\\
 Activation & NMSE$\downarrow$ & Acc.$\uparrow$ & VPT$\uparrow$ & Acc.$\uparrow$ & time ratio$\downarrow$ & ratio$\downarrow$\\
 \midrule
 MulClip $d=2$ & 0.087 & 0.981 & 473 & 0.771 & 0.34 & 0.94\\
 MulClip $d=3$ & 0.063 & 0.968 & 599 & 0.781 & 0.34 & 0.94\\
 Monotonic clip $d=2$ & 0.087 & 0.764 & 472 & 0.772 & --- & ---\\
 Monotonic clip $d=3$ & 0.063 & 0.781 & 582 & 0.785 & --- & ---\\
 \midrule
 \tanhfpu & 0.081 & 0.761 & 628 & 0.779 & 1.00 & 1.00\\
 Rational 3/2 & 0.104 & 0.767 & 596 & 0.779 & 0.40 & 0.94\\
 Rational 5/4 & 0.081 & 0.752 & 626 & 0.782 & 0.44 & 0.95\\
 LUT nearest 10--16384 & 0.252--0.110 & 0.697--0.990 & 54--391 & 0.679--0.776 & 0.39 & 0.95--5.53\\
 LUT interp.\ 16--256 & 0.144--0.098 & 0.763--0.746 & 435--469 & 0.757--0.779 & 0.43 & 0.95--1.02\\
 PWL8 & 0.153 & 0.777 & 363 & 0.756 & 0.37 & 0.95\\
 \midrule
 hardtanh & 0.150 & 0.803 & 390 & 0.742 & 0.32 & 0.94\\
 softsign & 0.127 & 0.734 & 470 & 0.748 & 0.33 & 0.94\\
 ReLU & 0.152 & 0.795 & 352 & 0.752 & 0.30 & 0.94\\
 Abe F1 / F2 / F3 \cite{abe} & \begin{tabular}[t]{@{}c@{}}0.266 / 0.604\\ / 0.142\end{tabular} & \begin{tabular}[t]{@{}c@{}}0.848 / 0.750\\ / 0.844\end{tabular} & \begin{tabular}[t]{@{}c@{}}253 / 5\\ / 311\end{tabular} & \begin{tabular}[t]{@{}c@{}}0.769 / 0.720\\ / 0.776\end{tabular} & \begin{tabular}[t]{@{}c@{}}0.34 / 0.48\\ / 0.40\end{tabular} & 0.95\\
 \bottomrule
 \end{tabular}

 \smallskip
 \parbox{\linewidth}{\footnotesize Note: $\downarrow$ lower is better; $\uparrow$ higher is better. Accuracy is the mean over 50 seeds. State-update time and flash usage are ratios to \tanhfpu\ (1.615~ms, 14,304~B). The monotonic-clip variants differ from MulClip only in the clipping-threshold constant, so their time and flash usage were not measured. For the LUT rows, the values for the smallest and largest table sizes are given in the order ``smallest-size value--largest-size value''; they are not ranges (all values in Appendix~\ref{app:C}). The state-update time ratios of the functions of Abe et al.\ are relative to \tanhfpu\ measured in the separate build that included them (Appendix~\ref{app:B}).}
\end{table}

Table~\ref{tab:main} shows the accuracy on the four tasks. On MG, \tanhfpu\ and the 5/4-order rational approximation slightly outperformed MulClip with $d=3$ (VPT 628 and 626 vs.\ 599). On KWS, the accuracies of all activations except the 10-entry LUT and Abe F2 fell within 0.742--0.785. In contrast, on Parity8, only MulClip and the nearest-neighbor LUTs with 100 or more entries reached 0.911--0.990, whereas all others were below 0.85 (0.761 for \tanhfpu). The interpolated LUTs, which have no steps, yielded 0.736--0.763, suggesting that quantization steps may have contributed to the nearest-neighbor LUT results. The monotonic functions of Abe et al.\ reached up to 0.848 on Parity8, but their NARMA-20 NMSEs were 0.142--0.604 (1.75--7.46 times that of \tanhfpu).

\subsection{Effect of Folding}\label{sec:folding}
On Parity8, MulClip raised the accuracy from 0.764 ($d=2$) and 0.781 ($d=3$) for the monotonic-clip variants to 0.981 and 0.968, respectively. Both use the same operations and differ only in the clipping-threshold constant. In this case, about 19\% and 17\% of the pre-activations fell in the folding regime, and about 11\% and 10\% in the zero regime. When the number of bits $k$ was varied from 2 to 10 (Appendix~\ref{app:E}), all activations achieved 0.99 or higher with no difference for $k\leq6$, whereas MulClip consistently outperformed the monotonic-clip variants for $k\geq7$. For $k\geq8$, MulClip also outperformed all monotonic activations other than the nearest-neighbor LUTs, but the 4096- and 16384-entry nearest-neighbor LUTs achieved accuracy comparable to or better than that of MulClip with $d=2$. To separate the contributions of the folding and zero regimes, we evaluated, for $d=2$, a variant with only part of the folding regime and a variant with only the zero regime (Appendix~\ref{app:D}). Their accuracies were 0.877 and 0.810, respectively, exceeding the monotonic-clip variant but falling short of MulClip, which has both regimes and achieved the highest accuracy.

As for the other tasks, on NARMA-20 the pre-activations never entered the folding regime, and the results of MulClip and the monotonic-clip variant were identical. On MG, the fraction entering it was 0.4\% or less, and the difference was small (Appendix~\ref{app:D}). On KWS with $d=3$, the search selected an operating point that uses the folding regime (9.8\% of the pre-activations), and the accuracy was slightly lower than that of the monotonic-clip variant (0.781 vs.\ 0.785). Because MulClip also reproduces 0.785 when given the hyperparameters of the monotonic-clip variant (Appendix~\ref{app:D}), that operating point exists within MulClip as well, but the 100-trial search did not select it. Thus, MulClip contains the operating points of the monotonic-clip variant and provides additional ones, but a finite search is not guaranteed to select the best operating point.

The low accuracy of the monotonic-clip variant could also stem from an inability to retain the most recent eight bits. However, when trained on the same input sequence to predict the number of 1s among the most recent eight bits, even the monotonic-clip variant predicted it almost without error (NMSE $\leq4\times10^{-8}$; Appendix~\ref{app:E}). Since parity is whether this count is even or odd, the difference on Parity8 is hard to explain by insufficient memory alone.

\section{Discussion and Conclusion}\label{sec:conclusion}
These results indicate that, for RC on MCUs, the design goal of an activation function need not be to approximate $\tanh$. Curvature was effective on NARMA-20, and the folding response was effective on Parity8.

We have not verified the mechanism by which the folding response helped on Parity8, so it remains a hypothesis. One hypothesis is that, when the pre-activation changes monotonically with the number of 1s, a folding response allows a single unit to raise and then lower its output, making it easier to represent parity, which alternates between 0 and 1 as the number of 1s increases; this corresponds to the fact that a nonmonotonic response can represent XOR with a single unit \cite{noel}.

This study has the following limitations. First, the state-update time was measured for a ring reservoir on a Cortex-M4. The share of the activation function is expected to decrease as the number of connections increases, but we did not measure the time for other topologies (Appendix~\ref{app:B}). In addition, the accuracies in Table~\ref{tab:main} were evaluated on a PC over 50 seeds, and on-device inference was confirmed only for NARMA-20 and Parity8 with five seeds. Second, because MulClip returns 0 for $|x|\geq c_d$, information may be lost for large inputs that do not appear during training. Furthermore, for $d=3$, the slope just below the threshold is $-2$, which locally amplifies the input. Third, $c_d$ is not the accuracy-maximizing threshold. For Parity8 with $d=2$, a threshold of $c=5$ yielded an accuracy of 0.999, but this setting creates a region in which the output sign is opposite to that of the input, and the output also falls outside $[-1,1]$. In contrast, $c_d$ is the threshold, determined by the functional form, that includes the folding regime to its full extent without causing sign inversion (Appendix~\ref{app:D}).

In summary, we proposed MulClip, an activation function composed only of comparison, absolute value, multiplication, and subtraction. On a Cortex-M4, MulClip ran with 0.34 times the state-update time of \tanhfpu, and in PC-based evaluation it achieved a lower NMSE on NARMA-20 than \tanhfpu\ and all 11 approximations, as well as higher accuracy on Parity8 than the monotonic-clip variant. As for the order, $d=3$, which yielded a clearly lower NMSE on NARMA-20 at a cost difference of only one cycle, is a reasonable default. For the evaluated ring reservoir, inexpensively realizing a response with both monotonic and folding regimes was more advantageous than approximating $\tanh$ in terms of both state-update time and task accuracy.

\section*{Acknowledgment}
The authors used the generative AI tool GitHub Copilot (GitHub Copilot CLI; model: Anthropic's Claude) to write code for the experimental firmware and for the PC-based experiment, analysis, and plotting scripts; to draft and revise the manuscript text; and to cross-check the numerical values in the text against the experimental data. The authors reviewed all code, numerical values, figures, and text and take full responsibility for the content.

\makeatletter
\renewcommand{\@biblabel}[1]{[#1]}
\makeatother

\clearpage
\appendix
\renewcommand{\thesection}{\Alph{section}}
\renewcommand{\thesubsection}{\thesection.\arabic{subsection}}
\makeatletter
\renewcommand{\p@subsection}{}
\makeatother
\numberwithin{table}{section}
\renewcommand{\thetable}{\thesection\arabic{table}}
\captionsetup[table]{textfont=normalfont}

\section*{Appendices}
\noindent The following appendices provide details of the definitions, experimental conditions, and results referred to in the main text. Reference and equation numbers ((1)--(4)) follow the main text. The terminology also follows the main text: $|x|\leq d/(d-1)$ is called the monotonic regime, $d/(d-1)<|x|<c_d$ the folding regime, and $|x|\geq c_d$ the zero regime. The control whose clipping threshold is placed at the extremum $d/(d-1)$ is called the monotonic-clip variant (``Monotonic clip'' in the tables).

\section{Implementation and Optimization of the Compared Activations}\label{app:A}
We give the definitions of the 20 activation functions compared in the main text. All were implemented in single-precision floating point (float32), and the same expressions and order of operations were used for the accuracy evaluation on the PC (Python) and the measurements on the MCU (C).

\paragraph{MulClip.}
As in (1)--(3) of the main text. For $d=2$, $f(x)=x_c-\tfrac14|x_c|x_c$, and for $d=3$, $f(x)=x_c-\tfrac{4}{27}x_c^3$, where $x_c$ is $x$ clipped to $[-c_d,c_d]$ ($c_2=4$, $c_3=3\sqrt3/2\approx2.598076$). The monotonic-clip variant, the control in Section~\mbox{III-D} of the main text, uses the same expression with the clipping threshold placed at the extremum $d/(d-1)$ (2 for $d=2$ and 1.5 for $d=3$). On the MCU, after clipping with two comparisons, the output is computed with an absolute value, two multiplications, and one subtraction for $d=2$, and with three multiplications and one subtraction for $d=3$. For $d=3$, the multiplications are performed in the order $(4/27)\cdot(x_c\cdot x_c)\cdot x_c$ to match the rounding order of the PC implementation. The clipping threshold appears only as a comparison constant, and its value does not change the type or number of instructions. The timing measurements and the check of agreement with the PC computation were also performed with $c_d$ (Appendices~\ref{app:B} and~\ref{app:F}).

\paragraph{\tanhfpu.}
newlib's \texttt{tanhf()} built with single-precision FPU instructions (\texttt{fpv4-sp-d16}).

\paragraph{Rational approximations.}
The 3/2-order approximation is $f(x)=x(27+x^2)/(27+9x^2)$, which has value 1 and slope 0 at $x=3$; for $|x|\geq3$, the input is clamped to $\pm3$. The 5/4-order approximation is $f(x)=x(945+105x^2+x^4)/(945+420x^2+15x^4)$, the Pad\'e approximant of $\tanh$; the input is clamped at $x\approx3.6467$, where the value reaches 1. In both cases the denominator is always positive, and one division is required.

\paragraph{LUTs.}
The values of $\tanh$ at $n$ points dividing $[-5,5]$ into $n-1$ equal intervals were stored as a constant array in flash. For nearest-neighbor lookup, the input is clamped to $[-5,5]$, the index is computed as $\mathrm{idx}=\lfloor (x_c+5)\cdot(n-1)/10+0.5\rfloor$, and the value at that point is returned ($n=10$, 100, 1000, 4096, 16384). For linear interpolation, the integer part $i$ and the fractional part of $t=(x_c+5)\cdot(n-1)/10$ are computed, and the values at the two adjacent points are linearly interpolated ($n=16$, 64, 256).

\paragraph{PWL8.}
A piecewise-linear approximation that divides $[-5,5]$ into eight segments and fits a straight line in each. The segment boundaries were placed to minimize the maximum error; the boundaries on the positive side are 0, 0.6726, 1.2955, 1.7690, and 5.0. The coefficients are odd-symmetric.

\paragraph{softsign, ReLU, and hardtanh.}
softsign is $x/(1+|x|)$ (one division), ReLU is $\max(x,0)$, and hardtanh clips $x$ to $[-1,1]$ (comparisons only).

\paragraph{Polynomial activation functions of Abe et al.\ (F1--F3) \cite{abe}.}
The output of the ninth-order polynomial
\[
 f(x)=\sum_{i=1}^{5}a_{2i-1}x^{2i-1}+\operatorname{sgn}(x)\sum_{i=1}^{4}a_{2i}x^{2i}
\]
is limited to $[-1,1]$, using the three coefficient sets of the original paper. For F1, $a_1$--$a_9$ are all 1; for F2, $(a_1,\ldots,a_9)=(1/256,1/256,1/64,1/64,1/16,1/16,1/4,1/4,1)$; and for F3, $(1,1/4,1/4,1/16,1/16,1/64,1/64,1/256,1/256)$. Because the polynomial $P(t)=\sum_k a_k t^k$ in $t=|x|$ has positive coefficients and increases monotonically for $t\geq0$, $|x|$ was clipped at the saturation point $t_{\mathrm{sat}}$ satisfying $P(t_{\mathrm{sat}})=1$ (0.5005 for F1, 0.9390 for F2, and 0.7321 for F3), and $\operatorname{sgn}(x)\cdot\min(P(t),1)$ was computed with Horner's method (on the MCU, the nine Horner stages were manually unrolled). The original work implemented these functions in fixed point on an FPGA and did not report their evaluation cost on an MCU.

\paragraph{Optimization.}
To avoid unfairly penalizing the approximate implementations, two optimizations were applied. First, the division in the LUT index computation was replaced by multiplication by the reciprocal $(n-1)/10$ determined at compile time (for the 1000-entry nearest-neighbor LUT, the state-update time decreased from 0.702~ms to 0.637~ms). The fraction of inputs whose index shifts by one owing to the difference in rounding was $2.1\times10^{-4}$ even for 16384 entries. Second, the segment search of PWL8 was changed from a binary search over the boundary array to a method that exploits odd symmetry and identifies one of the four segments on the $|x|$ side with at most three comparisons (the state-update time decreased from 1.092~ms to 0.592~ms). The mapping is identical before and after the optimization (except for differences of about 1 ULP due to rounding of the intercepts on the negative side).

\section{Details of the Measurements on the MCU}\label{app:B}
\subsection{Measurement environment}
An STM32 NUCLEO-F446RE (Cortex-M4, single-precision FPU) was operated at 84~MHz. The firmware was built with arm-none-eabi-gcc 13.3 with optimization \texttt{-O2} and hardware floating point (\texttt{-mfloat-abi=hard -mfpu=fpv4-sp-d16}). Measured values were sent to the PC via UART.

\subsection{Measurement method}
The 500-unit ring reservoir of (4) in the main text was run for 230 steps of scalar input. The input was the NARMA-20 input sequence (uniform random numbers in $[0,0.5]$). The state-update time per step was measured with the DWT cycle counter (CYCCNT) over the interval that includes computing the pre-activations of all units and evaluating the activation function. The number of cycles per activation was obtained by separately measuring only the evaluation of the activation function, using the actual pre-activation values of each step (including the function call). Interrupts were disabled during the measured intervals. All activations except those of Abe et al.\ were included in a single binary and measured in turn with common weights and inputs. The functions of Abe et al.\ were measured in a separate build that additionally included them, and their state-update time ratios were computed relative to \tanhfpu\ measured in the same build (note to Table~I of the main text). The readout (one output, a 500-dimensional inner product) was not included in the state-update time and was measured separately. The same binary was run four times, and the mean cycle count of each activation was exactly identical across the four runs.

\subsection{Flash usage}
Flash usage was obtained by building, for each activation, a binary that contains only that activation and taking the sum of \texttt{text} and \texttt{data} reported by arm-none-eabi-size. Owing to differences in the build environment, the increments have an uncertainty of up to about 8~B.

\begin{table}[H]
 \centering\small
 \caption{Measured values for all activations (500 units, per step)}\label{tab:S2}
 \begin{tabular}{lrrrr}
 \toprule
 Activation & Cycles/eval. & State-update time [ms] & Flash [B] & Increment from smallest [B]\\
 \midrule
 hardtanh & 25.0 & 0.513 & 13,440 & 16\\
 MulClip $d=2$ & 30.0 & 0.543 & 13,464 & 40\\
 MulClip $d=3$ & 31.0 & 0.549 & 13,480 & 56\\
 \tanhfpu & 203.2 & 1.615 & 14,304 & 880\\
 Rational 3/2 & 38.0 & 0.638 & 13,480 & 56\\
 Rational 5/4 & 49.0 & 0.704 & 13,520 & 96\\
 LUT nearest 10 & 44.5 & 0.629 & 13,544 & 120\\
 LUT nearest 100 & 44.6 & 0.630 & 13,904 & 480\\
 LUT nearest 1000 & 45.8 & 0.637 & 17,512 & 4,088\\
 LUT nearest 4096 & 44.0 & 0.626 & 29,880 & 16,456\\
 LUT nearest 16384 & 44.0 & 0.626 & 79,032 & 65,608\\
 LUT interp.\ 16 & 54.1 & 0.687 & 13,608 & 184\\
 LUT interp.\ 64 & 55.4 & 0.694 & 13,800 & 376\\
 LUT interp.\ 256 & 55.6 & 0.695 & 14,576 & 1,152\\
 PWL8 & 38.3 & 0.592 & 13,584 & 160\\
 softsign & 20.0 & 0.531 & 13,424 & 0\\
 ReLU & 21.0 & 0.490 & 13,424 & 0\\
 Abe F1 & 31.0 & 0.549 & 13,536 & 112\\
 Abe F2 & 70.0 & 0.781 & 13,552 & 128\\
 Abe F3 & 48.4 & 0.652 & 13,552 & 128\\
 \bottomrule
 \end{tabular}

 \smallskip
 \parbox{0.92\linewidth}{\footnotesize Note: The evaluation of \tanhfpu\ accounted for 74.9\% of the state-update time. The functions of Abe et al.\ were measured in a separate build (Appendix~B.2). The computation time of the readout (one output) was 0.051~ms per step.}
\end{table}

\subsection{Projection for reservoirs with denser connections (rough estimate)}
The measurements in the main text were performed with a ring topology (number of recurrent connections per unit $m=1$). We roughly estimated, from the measured values, the share of the activation function as $m$ increases. One step with \tanhfpu\ took 135,684 cycles, i.e., 271.4 cycles per unit, of which 203.2 cycles were for the activation function and 68.2 cycles for everything else. Assuming about 5 cycles per nonzero element of the sparse matrix (reading the value, column index, and state, and the multiply--accumulate), we added 5 cycles for each increment of $m$. Under this assumption, the share of \tanhfpu\ in the state-update time is about 64\% for $m=10$, about 39\% for $m=50$ (10\% connection density for $N=500$), and about 7\% for dense connections ($m=500$). Replacing \tanhfpu\ with MulClip with $d=3$ at $m=50$ is expected to reduce the state-update time by about 35\%. This is an unmeasured rough estimate, and the number of cycles per nonzero element depends on the data layout and implementation.

\section{Conditions and Results of the Accuracy Evaluation}\label{app:C}
\subsection{Reservoir}
The 500-unit ring reservoir of (4) in the main text was computed in float32. Training of the readout (ridge regression) and computation of its output were performed in float64. The coupling weight $w$ was searched as the spectral radius, and the input weights $v_i$ and biases $b_i$ were drawn from uniform distributions on $[-\text{input scale},\,\text{input scale}]$ and $[-\text{bias scale},\,\text{bias scale}]$, respectively. For KWS, 10-dimensional MFCC vectors were used as input, together with leaky integration with leak rate $a$, $x(t+1)=(1-a)x(t)+a f(\cdot)$.

\subsection{Tasks}
NARMA-20 is the task of estimating
\[
 y(t+1)=0.2\,y(t)+0.04\,y(t)\sum_{j=0}^{19}y(t-j)+1.5\,u(t)u(t-19)+0.001
\]
for a uniform random input $u(t)$ in $[0,0.5]$. We used 200 washout steps, 5,000 training steps, 1,500 validation steps, and 1,500 test steps, and the features consisted of the state augmented with a constant term and the input. The metric is the NMSE over the test segment. Parity8 is the task of determining the parity of the most recent eight bits of a uniform random bit sequence over $\{0,1\}$; a sequence was generated for each seed, with 4,000 training, 2,000 validation, and 2,000 test steps (the first 50 steps of each used as washout). The readout output was limited to $[0,1]$, and the task was evaluated by the accuracy of decisions with a threshold of 0.5. For MG, the standardized Mackey--Glass series with $\tau=17$ was used, with 200 washout, 5,000 training, 1,500 validation, and 2,000 test steps. After training for one-step-ahead prediction, the state was updated with the inputs of the validation segment, and then the test segment was predicted autonomously by feeding the predicted output back as the next input; the metric is the number of steps until the error first exceeds 0.5 times the standard deviation of the true values (VPT). For KWS, ten words of Google Speech Commands (yes, no, up, down, left, right, on, off, stop, go) were used with the official validation and test splits. The training data were limited to at most 300 samples per word. Ten MFCCs were extracted from each utterance (16~kHz, 1~s) with a hop length of 1,024, and classification was performed using the state at the final frame augmented with a constant term; the metric is accuracy.

\subsection{Hyperparameter search}
For each activation and each task, a 100-trial search was performed with Optuna's TPE \cite{optuna}. The objective of the search was the mean validation error over ten search seeds (1000--1009) (NMSE for NARMA-20, $1-\text{accuracy}$ for Parity8 and KWS, and the NMSE of the first 100 steps of autonomous prediction for MG). For NARMA-20 and MG, the search ranges were spectral radius $[0.1,1.2]$, input scale $[0.01,2.0]$, bias scale $[0,1]$, and ridge coefficient $[10^{-6},10^{-1}]$ (log-uniform). For Parity8 and the control experiments in Appendix~\ref{app:E}, because the selected input scale reached the upper bound, only the input-scale range was widened to $[0.01,20]$ (log-uniform). For KWS, the ranges were spectral radius $[0,2]$, input scale $[10^{-3},10]$, bias scale $[0,10]$, leak rate $[0.01,1]$, and ridge coefficient $[10^{-8},1]$ (log-uniform except for the spectral radius and bias scale). With the selected hyperparameters, the readout was retrained on the combined training and validation segments, and the test segment was evaluated with 50 seeds (2000--2049) separate from those used in the search. The seeds determine the reservoir weights (and, for Parity8, also the input sequence). The random seed of the search was common to all activations (1234).

\begin{table}[H]
 \centering\small
 \caption{Accuracy for each LUT size (mean over 50 seeds)}\label{tab:S3}
 \begin{tabular}{lcccc}
 \toprule
 Activation & NARMA-20 NMSE & Parity8 accuracy & MG VPT & KWS accuracy\\
 \midrule
 LUT nearest 10 & 0.2516 & 0.697 & 54 & 0.679\\
 LUT nearest 100 & 0.2092 & 0.911 & 100 & 0.746\\
 LUT nearest 1000 & 0.1558 & 0.980 & 208 & 0.771\\
 LUT nearest 4096 & 0.1310 & 0.979 & 301 & 0.772\\
 LUT nearest 16384 & 0.1104 & 0.990 & 391 & 0.776\\
 LUT interp.\ 16 & 0.1439 & 0.763 & 435 & 0.757\\
 LUT interp.\ 64 & 0.1271 & 0.736 & 401 & 0.778\\
 LUT interp.\ 256 & 0.0984 & 0.746 & 469 & 0.779\\
 \bottomrule
 \end{tabular}

 \smallskip
 \parbox{0.85\linewidth}{\footnotesize Note: Table~I of the main text shows only the values for the smallest and largest table sizes. The accuracies of the other activations are as given in Table~I of the main text.}
\end{table}

\section{Details of the Effect of Folding}\label{app:D}
\subsection{Difference in accuracy with and without folding}\label{app:D.1}
MulClip (clipping threshold $c_d$) and the monotonic-clip variant (clipping threshold $d/(d-1)$) differ only in the value of the clipping threshold, and the hyperparameters of each were searched with the same procedure as in Appendix~\ref{app:C}. The accuracies of both are shown in Table~I of the main text. On Parity8, folding improved the accuracy by 0.217 for $d=2$ and by 0.187 for $d=3$. On NARMA-20, the results of the two, including the selected hyperparameters, were exactly identical. The differences in MG VPT were 1 step for $d=2$ and 17 steps for $d=3$. On KWS with $d=3$, MulClip was lower by 0.0043; the origin of this difference is examined in Appendix~\ref{app:D.3}.

\subsection{Fractions of pre-activations in each regime}\label{app:D.2}
For MulClip, the 50-seed evaluation was rerun with the selected hyperparameters, and the pre-activation values input to the activation function were recorded for all steps and all units in the training and evaluation segments (including washout). Table~\ref{tab:S4a} shows the fraction of pre-activations whose absolute value exceeded the extremum $d/(d-1)$ (the sum of the folding and zero regimes) and its breakdown.

\begin{table}[H]
 \centering\small
 \caption{Distribution of the pre-activations of MulClip (50 seeds, all steps and all units in training and evaluation)}\label{tab:S4a}
 \begin{tabular}{lcccc}
 \toprule
 & & Beyond the extremum & Folding regime & Zero regime\\
 Task & $d$ & ($|x|>d/(d-1)$) & ($d/(d-1)<|x|<c_d$) & ($|x|\geq c_d$)\\
 \midrule
 NARMA-20 & 2 & 0\% & 0\% & 0\%\\
 NARMA-20 & 3 & 0\% & 0\% & 0\%\\
 Parity8 & 2 & 30.1\% & 19.3\% & 10.8\%\\
 Parity8 & 3 & 26.8\% & 16.6\% & 10.2\%\\
 MG & 2 & 0\% & 0\% & 0\%\\
 MG & 3 & 0.36\% & 0.36\% & 0\%\\
 KWS & 2 & 0.0001\% & 0.0001\% & 0\%\\
 KWS & 3 & 9.8\% & 9.8\% & 0\%\\
 \bottomrule
 \end{tabular}

 \smallskip
 \parbox{0.85\linewidth}{\footnotesize Note: The extremum $d/(d-1)$ is 2 for $d=2$ and 1.5 for $d=3$, and $c_d$ is 4 for $d=2$ and about 2.60 for $d=3$. The fraction in the folding regime is the fraction beyond the extremum minus the fraction in the zero regime. 0\% indicates that none of the recorded values satisfied the condition.}
\end{table}

On NARMA-20, the pre-activations never entered either the folding regime or the zero regime. In this case, the output of MulClip is identical to that of the monotonic-clip variant, which is consistent with the exactly identical results of the two in Appendix~\ref{app:D.1}. On Parity8, 27--30\% of the pre-activations exceeded the extremum. The breakdown is 19.3\% ($d=2$) and 16.6\% ($d=3$) of all pre-activations in the folding regime, and 10.8\% and 10.2\% in the zero regime. The contributions of the two regimes are separated in Appendix~\ref{app:D.4}. On MG, the fraction beyond the extremum was 0.4\% or less, and on KWS with $d=3$, 9.8\% entered the folding regime. Which regime is used is determined by the set of hyperparameters selected by the search.

\subsection{Verification by swapping hyperparameters}\label{app:D.3}
To determine whether the difference on KWS ($d=3$) in Appendix~\ref{app:D.1} stems from the difference in functions or from the difference in the operating points reached by the searches, the hyperparameters selected for the monotonic-clip variant were given to MulClip and evaluated with the same 50 seeds. The mean accuracy was 0.785, which agreed with the value of the monotonic-clip variant itself (0.785) to the sixth decimal place. That is, the operating point of the monotonic-clip variant is also contained in MulClip, and the difference in Appendix~\ref{app:D.1} arises because the search for MulClip reached a different operating point that uses the folding regime (9.8\% in Appendix~\ref{app:D.2}). Conversely, when the hyperparameters selected for MulClip were given to the monotonic-clip variant, the accuracy decreased to 0.769.

\subsection{Sensitivity analysis of the clipping threshold and separation of the folding and zero regimes}\label{app:D.4}
In the main text, the clipping threshold is fixed at $c_d$ and is not searched for each task. Here, for reference, we show the accuracy when the clipping threshold is varied around $c_d$. This is intended to check the validity of the fixed $c_d$ post hoc, and all results in the main text use $c_d$. For each clipping threshold, the hyperparameters were searched again with the same procedure as in Appendix~\ref{app:C}.

\begin{table}[H]
 \centering\small
 \caption{Accuracy on Parity8 ($d=2$) for different clipping thresholds $c$ and for the zero-regime-only variant (mean over 50 seeds)}\label{tab:S4b}
 \begin{tabular}{clc}
 \toprule
 $c$ & Response shape & Accuracy\\
 \midrule
 2.0 & Monotonic (monotonic-clip variant) & 0.764\\
 3.0 & Folding stops partway & 0.877\\
 4.0 & $c_d$ (MulClip in the main text) & 0.981\\
 5.0 & Beyond $c_d$ (with a sign-inverted region) & 0.999\\
 4.0 & Zero regime only (constant 1 from the extremum to $c_d$, 0 outside) & 0.810\\
 \bottomrule
 \end{tabular}
\end{table}

On Parity8, as the clipping threshold was widened from the extremum (2.0 for $d=2$) to $c_d$, the accuracy increased from 0.764 to 0.877 and then to 0.981. This indicates that a wider folding range is more advantageous for Parity8, consistent with the results in Section~\mbox{III-D} of the main text.

\paragraph{Separation of the folding and zero regimes.}
For Section~\mbox{III-D} of the main text, a variant with only the zero regime was also evaluated with the same procedure (Appendix~\ref{app:C}). This variant returns the same response as the monotonic-clip variant for $|x|\leq d/(d-1)$, a constant $\pm1$ for $d/(d-1)<|x|<c_d$, and 0 for $|x|\geq c_d$, and has no folding regime. On the other hand, the variant with $c=3$ (the ``variant with only part of the folding regime'' in the main text) has no zero regime, and its output is constant at 0.75 for $|x|>3$. Their accuracies were 0.810 and 0.877, respectively, exceeding the monotonic-clip variant (0.764) but falling short of MulClip (0.981). Hence, the folding and zero regimes each raise the Parity8 accuracy on their own, but the accuracy is highest with the MulClip response, which has both.

With $c=5.0$, beyond $c_d$, the accuracy was even higher at 0.999. Therefore, $c_d$ is not the value that maximizes the accuracy on Parity8. We nevertheless use $c_d$ because $c_d$ is determined not by accuracy but by a property of the response shape. For $c>c_d$, the sign of the output is opposite to that of the input in the range where $|x|$ exceeds $c_d$. Furthermore, the absolute value of the output exceeds 1 for $c>2+2\sqrt2\approx4.83$ when $d=2$ (for $c=5.0$, $f(\pm5)=\mp1.25$), and likewise for $c>3$ when $d=3$. That is, $c_d$ is the largest clipping threshold for which the sign of the output agrees with that of the input, and the property stated in Section~II of the main text, ``the output always lies in $[-1,1]$,'' also holds at $c_d$. Pursuing accuracy beyond $c_d$ would mean returning to the practice of searching the clipping threshold for each task, which is outside the scope of the claims of the main text.

\section{Control Experiments (Count8, Parity-$k$)}\label{app:E}
\subsection{Memory-length control (Count8)}\label{app:E.1}
We give the details of the task described in Section~\mbox{III-D} of the main text, predicting the number of 1s in the most recent eight bits (hereafter Count8). The same input sequence as Parity8 (uniform random bit sequence over $\{0,1\}$ generated with the same seeds), data lengths, washout, readout, and search procedure were used, and only the target was changed. The target is $s/8$, the number $s$ of 1s in the most recent eight bits divided by 8, evaluated by NMSE. The target of Parity8 is the parity of $s$.

\begin{table}[H]
 \centering\small
 \caption{NMSE on Count8 (mean over 50 seeds)}\label{tab:S5a}
 \begin{tabular}{lc}
 \toprule
 Activation & Count8 NMSE\\
 \midrule
 MulClip $d=2$ & $3.31\times10^{-8}$\\
 MulClip $d=3$ & $1.12\times10^{-8}$\\
 Monotonic clip $d=2$ & $3.31\times10^{-8}$\\
 Monotonic clip $d=3$ & $1.12\times10^{-8}$\\
 \tanhfpu & $1.06\times10^{-8}$\\
 \bottomrule
 \end{tabular}
\end{table}

All activations, including the monotonic-clip variants, reached an NMSE of $4\times10^{-8}$ or less on Count8. Therefore, the number of 1s in the most recent eight bits can be linearly read out from the state of the 500-unit ring reservoir, and the low accuracy of the monotonic-clip variant on Parity8 is hard to explain by insufficient memory length alone. However, we have not verified whether the information of each bit required for parity is retained, nor the contributions of individual units.

\subsection{Parity with different numbers of bits (Parity-$k$)}\label{app:E.2}
To check that the Parity8 results are not specific to eight bits, we evaluated Parity-$k$ ($k=2,3,4,5,6,7,9,10$), whose target is the parity of the number of 1s in the most recent $k$ bits. Apart from $k$, the conditions were the same as for Parity8, and the hyperparameters were searched again for each activation at each $k$. For $k\leq6$, the accuracies of all 22 configurations were 0.994 or higher, with no difference among activations; Table~\ref{tab:S5b} therefore shows the results for $k\geq7$. The $k=8$ column contains the values of Table~I of the main text and Table~\ref{tab:S3}.

\begin{table}[H]
 \centering\small
 \caption{Accuracy on Parity-$k$ ($k\geq7$) (mean over 50 seeds)}\label{tab:S5b}
 \begin{tabular}{lcccc}
 \toprule
 Activation & $k=7$ & $k=8$ & $k=9$ & $k=10$\\
 \midrule
 MulClip $d=2$ & 1.000 & 0.981 & 0.802 & 0.637\\
 MulClip $d=3$ & 1.000 & 0.968 & 0.754 & 0.625\\
 Monotonic clip $d=2$ & 0.972 & 0.764 & 0.625 & 0.559\\
 Monotonic clip $d=3$ & 0.960 & 0.781 & 0.637 & 0.568\\
 \tanhfpu & 0.937 & 0.761 & 0.622 & 0.561\\
 Rational 3/2 & 0.949 & 0.767 & 0.624 & 0.564\\
 Rational 5/4 & 0.935 & 0.752 & 0.620 & 0.562\\
 LUT nearest 10 & 0.891 & 0.697 & 0.600 & 0.544\\
 LUT nearest 100 & 0.996 & 0.911 & 0.709 & 0.579\\
 LUT nearest 1000 & 1.000 & 0.980 & 0.774 & 0.599\\
 LUT nearest 4096 & 1.000 & 0.979 & 0.796 & 0.647\\
 LUT nearest 16384 & 1.000 & 0.990 & 0.798 & 0.660\\
 LUT interp.\ 16 & 0.988 & 0.763 & 0.623 & 0.561\\
 LUT interp.\ 64 & 0.962 & 0.736 & 0.620 & 0.561\\
 LUT interp.\ 256 & 0.944 & 0.746 & 0.619 & 0.562\\
 PWL8 & 0.985 & 0.777 & 0.624 & 0.561\\
 hardtanh & 0.997 & 0.803 & 0.642 & 0.566\\
 softsign & 0.914 & 0.734 & 0.602 & 0.546\\
 ReLU & 0.997 & 0.795 & 0.598 & 0.531\\
 Abe F1 & 1.000 & 0.848 & 0.651 & 0.561\\
 Abe F2 & 0.958 & 0.750 & 0.605 & 0.544\\
 Abe F3 & 0.998 & 0.844 & 0.653 & 0.568\\
 \bottomrule
 \end{tabular}
\end{table}

For $k\geq7$, the accuracy decreased as $k$ increased, and the following trends were observed. First, MulClip outperformed the monotonic-clip variant of the same $d$ for all $k=7$--10 (the differences were 0.028--0.217 for $d=2$ and 0.040--0.187 for $d=3$). Second, for $k=8$--10, MulClip outperformed all monotonic activations other than the nearest-neighbor LUTs. Third, the 4096- and 16384-entry nearest-neighbor LUTs were comparable to or better than MulClip with $d=2$. Hence, MulClip's advantage over the monotonic-clip variant is not specific to $k=8$ and was observed for all of $k=7$--10, where the decision becomes difficult.

\section{Agreement Between the PC and the Device, and On-Device Inference}\label{app:F}
\subsection{Agreement of states}\label{app:F.1}
Because the accuracies in the main text were obtained by float32 simulation on the PC, we confirmed, for four representative activations (hardtanh, MulClip $d=2$, MulClip $d=3$, and \tanhfpu), that the computation on the device agrees with that on the PC. For this check, we used the same 500-unit ring reservoir as in Appendix~\ref{app:B}, the same weights as the PC implementation, and the same input (230 steps of the NARMA-20 input sequence). On the device, the states of all units at steps 0--4 and 200--229 were output via UART as raw floating-point bit patterns (in hexadecimal) and compared with the states computed under the same conditions on the PC ($500$ units $\times$ 35 steps $=$ 17,500 elements).

For hardtanh, the states agreed bitwise for all compared elements. Because hardtanh consists only of comparisons, this agreement shows that the weight layout and the state-update expression are identical to those of the PC implementation. For \tanhfpu, the maximum absolute error was $7.0\times10^{-3}$. This difference results from differences of a few ULPs, due to the different algorithms of newlib's \texttt{tanhf} and $\tanh$ on the PC (NumPy), accumulating through the recurrent connections.

MulClip was checked with the same $c_d$ as in the main text. Because the pre-activations do not reach the folding regime with the weights of Appendix~\ref{app:B}, only the input scale was changed to 10.0 so that the pre-activations entered both the folding and zero regimes (folding regime: 24.5\% for $d=2$ and 35.0\% for $d=3$; zero regime: 2.5\% and 14.7\%, respectively). The maximum absolute error relative to the float32 computation on the PC was $5.2\times10^{-7}$ for $d=2$ and $2.0\times10^{-6}$ for $d=3$. These differences arise because the compiler (\texttt{-O2}) combined the multiply--accumulate and the final subtraction of MulClip into fused multiply--add (FMA) instructions, reducing the number of roundings by one. When these fusions were emulated on the PC, the states agreed bitwise for all 17,500 elements for both $d=2$ and $d=3$.

For the approximate implementations in Appendix~\ref{app:A}, the C code of the firmware was built on the PC, and we confirmed that the outputs for 20,001 input points agreed with those of the PC implementation (except for index shifts due to the optimization in Appendix~\ref{app:A} and differences within $6\times10^{-8}$).

\subsection{On-device inference (NARMA-20 and Parity8)}\label{app:F.2}
In addition to the agreement of states, for the two main tasks, inference over the evaluation segment was performed on the device and the metrics were obtained directly. The readout was trained on the PC with the same procedure as in the main text (same hyperparameters and training segment), and its coefficients, together with the input weights, biases, and inputs of the evaluation segment, were placed in flash (about 260~KB). On the device, the state update and readout over the evaluation segment (for NARMA-20, 1,500 of the 1,700 steps excluding the first 200; for Parity8, 1,950 of the 2,000 steps excluding the first 50) were executed in float32, and the predicted values were output via UART as raw bit patterns. The metrics were computed on the PC from the same target values. The configurations were eight in total: \tanhfpu\ and MulClip ($d=2,3$) for NARMA-20, and these plus the monotonic-clip variants ($d=2,3$) for Parity8, using five of the evaluation seeds (2000--2004).

\begin{table}[H]
 \centering\small
 \caption{Comparison of on-device inference and PC computation (mean over five seeds)}\label{tab:S6}
 \begin{tabular}{llcc}
 \toprule
 Task & Activation & Device & PC\\
 \midrule
 NARMA-20 (NMSE) & \tanhfpu & 0.08340 & 0.08340\\
 NARMA-20 (NMSE) & MulClip $d=2$ & 0.08999 & 0.08999\\
 NARMA-20 (NMSE) & MulClip $d=3$ & 0.06472 & 0.06472\\
 Parity8 (accuracy) & \tanhfpu & 0.7483 & 0.7487\\
 Parity8 (accuracy) & Monotonic clip $d=2$ & 0.7535 & 0.7518\\
 Parity8 (accuracy) & Monotonic clip $d=3$ & 0.7588 & 0.7589\\
 Parity8 (accuracy) & MulClip $d=2$ & 0.9825 & 0.9825\\
 Parity8 (accuracy) & MulClip $d=3$ & 0.9670 & 0.9670\\
 \bottomrule
 \end{tabular}

 \smallskip
 \parbox{0.75\linewidth}{\footnotesize Note: The PC values are those of the procedure in the main text, with the readout computed in float64. Because they are means over five seeds, they differ from the 50-seed means in Table~I of the main text.}
\end{table}

The differences between the device and PC values arise because the device computes the readout in float32 and because of slight differences in the states (the fused multiply--add and the \texttt{tanhf} differences in Appendix~\ref{app:F.1}). Accordingly, the main accuracy comparisons in the main text (the advantage of $d=3$ on NARMA-20 and the effect of folding on Parity8) were reproduced with nearly the same values by on-device inference. However, only five seeds were checked, and MG and KWS were not run on the device.

\begin{thebibliography}{99}
\small
\bibitem{warden} P. Warden and D. Situnayake, \emph{TinyML: Machine Learning with TensorFlow Lite on Arduino and Ultra-Low-Power Microcontrollers}. Sebastopol, CA, USA: O'Reilly Media, 2019.
\bibitem{mcunet} J. Lin, W.-M. Chen, Y. Lin, J. Cohn, C. Gan, and S. Han, ``MCUNet: Tiny deep learning on IoT devices,'' in \emph{Proc. NeurIPS}, 2020, pp. 11711--11722.
\bibitem{cmsis} L. Lai, N. Suda, and V. Chandra, ``CMSIS-NN: Efficient neural network kernels for Arm Cortex-M CPUs,'' arXiv:1801.06601, 2018.
\bibitem{jaeger} H. Jaeger, ``The `echo state' approach to analysing and training recurrent neural networks,'' GMD---German National Research Center for Information Technology, Bonn, Germany, GMD Rep. 148, 2001.
\bibitem{lukosevicius} M. Luko\v{s}evi\v{c}ius and H. Jaeger, ``Reservoir computing approaches to recurrent neural network training,'' \emph{Comput. Sci. Rev.}, vol. 3, no. 3, pp. 127--149, 2009.
\bibitem{rodan} A. Rodan and P. Ti\v{n}o, ``Minimum complexity echo state network,'' \emph{IEEE Trans. Neural Netw.}, vol. 22, no. 1, pp. 131--144, Jan. 2011.
\bibitem{arm} ARM Ltd., \emph{Cortex-M4 Technical Reference Manual}, ARM DDI 0439B, 2010.
\bibitem{amin} H. Amin, K. M. Curtis, and B. R. Hayes-Gill, ``Piecewise linear approximation applied to nonlinear function of a neural network,'' \emph{IEE Proc.---Circuits Devices Syst.}, vol. 144, no. 6, pp. 313--317, 1997.
\bibitem{baker} G. A. Baker, Jr. and P. Graves-Morris, \emph{Pad\'e Approximants}, 2nd ed. Cambridge, U.K.: Cambridge Univ. Press, 1996.
\bibitem{glorot} X. Glorot and Y. Bengio, ``Understanding the difficulty of training deep feedforward neural networks,'' in \emph{Proc. AISTATS}, 2010, pp. 249--256.
\bibitem{nair} V. Nair and G. E. Hinton, ``Rectified linear units improve restricted Boltzmann machines,'' in \emph{Proc. ICML}, 2010, pp. 807--814.
\bibitem{hurley} L. A. Hurley, J. G. Restrepo, and S. E. Shaheen, ``Tuning the activation function to optimize the forecast horizon of a reservoir computer,'' \emph{J. Phys. Complexity}, vol. 5, no. 3, Art. no. 035004, Jul. 2024, doi: \href{https://doi.org/10.1088/2632-072X/ad5e55}{10.1088/2632-072X/ad5e55}.
\bibitem{noel} M. M. Noel, A. L, A. Trivedi, and P. Dutta, ``Growing cosine unit: A novel oscillatory activation function that can speedup training and reduce parameters in convolutional neural networks,'' arXiv:2108.12943, 2021.
\bibitem{smith} J. O. Smith III, \emph{Physical Audio Signal Processing for Virtual Musical Instruments and Audio Effects}. W3K Publishing, 2010. [Online]. Available: \url{https://ccrma.stanford.edu/~jos/pasp/} (accessed Oct. 26, 2026).
\bibitem{abe} Y. Abe, K. Nishida, M. Akai-Kasaya, and T. Asai, ``High-order polynomial activation function and regenerative internal weights for FPGA implementation of reservoir computing,'' \emph{Nonlinear Theory Appl., IEICE}, vol. 15, no. 2, pp. 262--272, Apr. 2024, doi: \href{https://doi.org/10.1587/nolta.15.262}{10.1587/nolta.15.262}.
\bibitem{yoshioka} K. Yoshioka and H. Tamukoh, ``A 340-\textmu{}W TinyML using LUT-based reservoir computing on low-cost FPGAs,'' \emph{IEEE Embedded Syst. Lett.}, vol. 17, no. 5, pp. 357--360, 2025, doi: \href{https://doi.org/10.1109/LES.2025.3598209}{10.1109/LES.2025.3598209}.
\bibitem{atiya} A. F. Atiya and A. G. Parlos, ``New results on recurrent network training: Unifying the algorithms and accelerating convergence,'' \emph{IEEE Trans. Neural Netw.}, vol. 11, no. 3, pp. 697--709, May 2000.
\bibitem{bertschinger} N. Bertschinger and T. Natschl\"ager, ``Real-time computation at the edge of chaos in recurrent neural networks,'' \emph{Neural Comput.}, vol. 16, no. 7, pp. 1413--1436, 2004.
\bibitem{mackey} M. C. Mackey and L. Glass, ``Oscillation and chaos in physiological control systems,'' \emph{Science}, vol. 197, no. 4300, pp. 287--289, 1977.
\bibitem{speech} P. Warden, ``Speech Commands: A dataset for limited-vocabulary speech recognition,'' arXiv:1804.03209, 2018.
\bibitem{optuna} T. Akiba, S. Sano, T. Yanase, T. Ohta, and M. Koyama, ``Optuna: A next-generation hyperparameter optimization framework,'' in \emph{Proc. ACM SIGKDD}, 2019, pp. 2623--2631.
\end{thebibliography}
\end{document}